\documentclass[sigconf]{acmart}
\renewcommand\footnotetextcopyrightpermission[1]{} % removes footnote with conference information in first column
\AtBeginDocument{%
  \providecommand\BibTeX{{%
    \normalfont B\kern-0.5em{\scshape i\kern-0.25em b}\kern-0.8em\TeX}}}

\setcopyright{acmlicensed}
\copyrightyear{2026}
\acmYear{2026}
\acmDOI{XXXXXXX.XXXXXXX}

\acmConference[WSDM '26]{The 19th ACM International Conference on Web Search and Data Mining}{February 22--26, 2026}{Boise, USA} 
\acmBooktitle{WSDM '26: The 19th ACM International Conference on Web Search and Data Mining,
 February 22--26, 2026, Boise, USA} 
\acmISBN{978-1-4503-XXXX-X/18/06}   % 待更新

\usepackage{textcomp}
\usepackage{verbatim}
\usepackage{amsfonts}
\usepackage{amsmath}
\usepackage{array}
\usepackage{subfigure}
\usepackage{caption}
\usepackage{balance}
\usepackage{multicol}
\usepackage{multirow}
\usepackage{color}
\usepackage{epsfig}
\usepackage{natbib}
\usepackage{xspace}
\usepackage{booktabs}
\usepackage{bm}
\usepackage{enumitem}
\usepackage{diagbox}
\usepackage{url}
\usepackage{geometry}
\usepackage{changepage}
\usepackage{textcomp}
\usepackage{graphicx}
\usepackage{color}
\usepackage{bm}
  
\usepackage{amssymb}
\usepackage{algorithm}
\usepackage{algorithmic}
\usepackage{threeparttable}
\usepackage{mathrsfs}

\newcommand{\model}{\textit{MOON}\xspace}
\newcommand{\bench}{\textit{MBE}\xspace}

\newcommand{\vpara}[1]{\vspace{0.05in}\noindent\textbf{#1 }}

\begin{document}

\title{MOON: Generative MLLM-based Multimodal Representation Learning for E-commerce Product Understanding}

% \author{Daoze Zhang$^{*}$, Zhanheng Nie$^{*}$, Jianyu Liu$^{*}$, Chenghan Fu$^{*\dagger}$, Wanxian Guan \\ Yuan Gao, Jun Song, Pengjie Wang, Jian Xu and Bo Zheng$^{\ddagger}$ }
% \affiliation{%
%   \institution{Alibaba Group}
%   \city{Hangzhou}
%   \country{China}
% }
% \email{ 
% {zhangdaoze.zdz, niezhanheng.nzh, liuyu.ljy, fuchenghan.fch, wanxian.gwx}@taobao.com,
% }
% \email{ 
% {gy275873, jsong.sj, pengjie.wpj, xiyu.xj}@taobao.com, 
% bozheng@alibaba-inc.com 
% }

\author{Daoze Zhang$^{*}$}
\affiliation{ \institution{Alibaba Group}\city{Hangzhou}\country{China} }
\email{ zhangdaoze.zdz@taobao.com }

\author{Chenghan Fu$^{*}$$^{\dagger}$}
\affiliation{ \institution{Alibaba Group}\city{Hangzhou}\country{China} }
\email{ fuchenghan.fch@taobao.com }

\author{Zhanheng Nie$^{*}$}
\affiliation{ \institution{Alibaba Group}\city{Hangzhou}\country{China} }
\email{ niezhanheng.nzh@taobao.com }

\author{Jianyu Liu$^{*}$}
\affiliation{ \institution{Alibaba Group}\city{Hangzhou}\country{China} }
\email{ liuyu.ljy@taobao.com }

\author{Wanxian Guan}
\affiliation{ \institution{Alibaba Group}\city{Hangzhou}\country{China} }
\email{ wanxian.gwx@taobao.com }

\author{Yuan Gao}
\affiliation{ \institution{Alibaba Group}\city{Hangzhou}\country{China} }
\email{ gy275873@taobao.com }

\author{Jun Song}
\affiliation{ \institution{Alibaba Group}\city{Hangzhou}\country{China} }
\email{ jsong.sj@taobao.com }

\author{Pengjie Wang}
\affiliation{ \institution{Alibaba Group}\city{Hangzhou}\country{China} }
\email{ pengjie.wpj@taobao.com }

\author{Jian Xu}
\affiliation{ \institution{Alibaba Group}\city{Hangzhou}\country{China} }
\email{ xiyu.xj@taobao.com }

\author{Bo Zheng$^{\ddagger}$}
\affiliation{ \institution{Alibaba Group}\city{Hangzhou}\country{China} }
\email{ bozheng@alibaba-inc.com  }

\thanks{$^{*}$ Equal Contribution. }
\thanks{$^{\dagger}$ Project Leader. }
\thanks{$^{\ddagger}$ Corresponding Author. }

\renewcommand{\shortauthors}{Daoze Zhang et al.}

%   <concept_desc>Applied computing~Health care information systems</concept_desc>

\ccsdesc[500]{Information systems~Recommender systems}

\keywords{Multi-modal Representations, Product Understanding, Recommendation System, E-commerce Search}



\begin{abstract}

With the rapid advancement of e-commerce, 
exploring general representations rather than task-specific ones 
% has become increasingly important, 
has attracted increasing research attention.
% as it enhances model generalizability across diverse downstream tasks while reducing training and maintenance costs.
% For product understanding, existing discriminative dual-encoder architectures are typically trained with instance-level contrastive learning or masked modeling objectives,
% which not only neglect valuable inter-item correlations implied in positive user feedback, but also inherently struggle to model the many-to-one alignment between product images and text.
For product understanding, although existing discriminative dual-flow architectures drive progress in this field, they inherently struggle to model the many-to-one alignment between multiple images and texts of products.
Therefore, we argue that generative Multimodal Large Language Models (MLLMs) hold significant potential for improving product representation learning. 
Nevertheless, achieving this goal still remains non-trivial due to several key challenges: 
the lack of multimodal and aspect-aware modeling modules in typical LLMs; 
the common presence of background noise in product images; 
and the absence of a standard benchmark for evaluation.
To address these issues, we propose the first generative MLLM-based model named \model for product representation learning. 
Our method 
(1) employs a guided Mixture-of-Experts (MoE) module for targeted modeling of multimodal and aspect-specific product content;
(2) 
% adopts MLLM’s visual grounding ability to 
effectively detects core semantic regions in product images to 
mitigate the distraction and interference caused by background noise;
% suppressing irrelevant background noise; 
and
(3) introduces the specialized negative sampling strategy to increase the difficulty and diversity of negative samples.
In addition, we release a large-scale multimodal benchmark \bench for various product understanding tasks. 
Experimentally, our model demonstrates competitive zero-shot performance on both our benchmark and the public dataset, showcasing strong generalization across various downstream tasks, including cross-modal retrieval, product classification, and attribute prediction.
% Furthermore, qualitative analysis and case studies further validate the effectiveness of \model in capturing nuanced semantic understanding of products in real-world e-commerce settings.
Furthermore, the case study and visualization illustrate the effectiveness of \model for product understanding. % in real-world e-commerce scenarios.
% The homepage of our work is in \url{https://anonymous.4open.science/r/product/}. 
The data of our \bench benchmark is given in \url{https://huggingface.co/datasets/Daoze/MM-Bench-E-Commerce}.

\end{abstract}

\maketitle

\section{Introduction}\label{sec:intro}

With the rapid growth of e-commerce, people's daily lives have become increasingly dependent on the convenience of online shopping. This trend has catalyzed a wide range of application tasks related to \textit{product understanding}, such as product retrieval~\citep{zhan2021product1m,dong2022m5product} and product classification~\citep{pawlowski2022machine,xu2019open}. 
Instead of designing specific models and algorithms for each specific task, exploring methods to learn general representations that can support various downstream tasks in the e-commerce domain offers significant advantages, which not only enhance the model generalizability but also greatly reduce the training and maintenance costs.

For the product understanding in the e-commerce field, earlier researches predominantly focus on extracting sparse ID-based features for products. 
These ID-based features are typically learned using collaborative filtering (CF) signals~\citep{ricci2010introduction} rather than the visual and textual \textit{content} (such as the image and title) of the products themselves. 
This not only leads to a gap between the representation learning process and the semantics of the product, but also results in the weakness or failure of the ID-based features in long-tail or cold-start scenarios where collaborative signals are sparse~\citep{yuan2020parameter}.

\begin{figure}[t]
  \centering
  \includegraphics[width=\linewidth / 100 * 90]{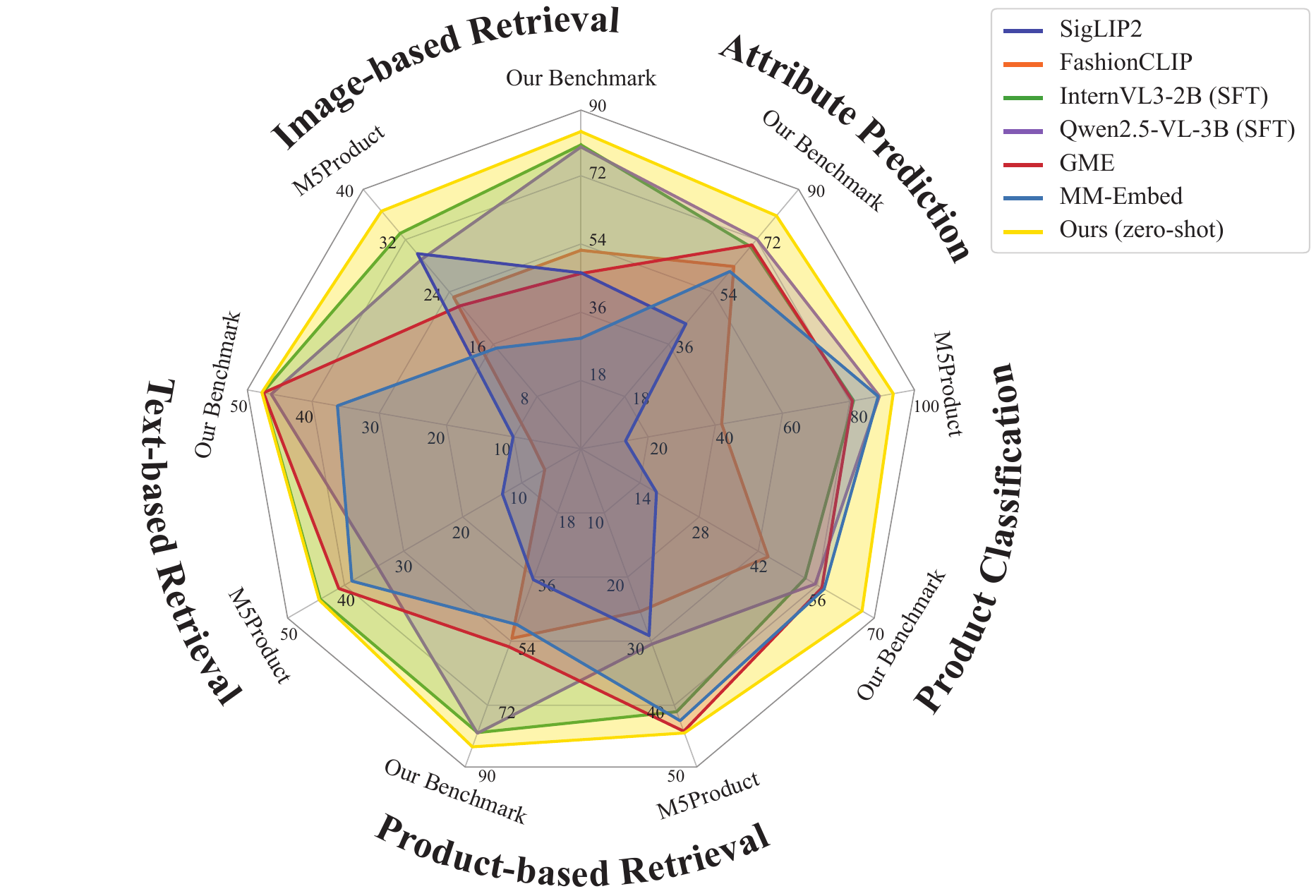}

  % \vspace{-2mm}
  
  \caption{Overall results on all the downstream tasks.}
  \label{fig:radar}

  % \vspace{-5mm}
\end{figure}

\begin{figure}[t]
  \centering
  \includegraphics[width=\linewidth / 100 * 90]{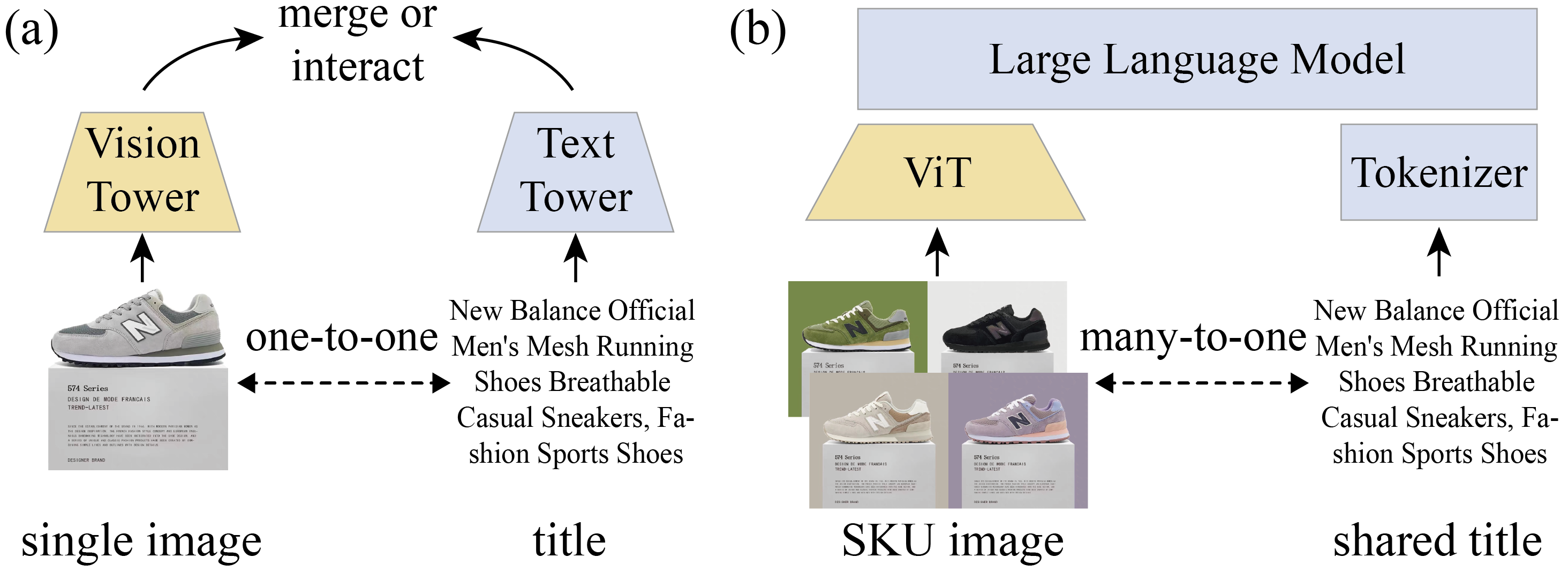}

  % \vspace{-3mm}
  
  \caption{Comparison between the dual-flow and MLLM architectures. 
  \small 
  (a) The dual-flow paradigm is inherently limited to encode one-to-one image-text pairs and cannot directly capture many-to-one relationships. 
  (b) The MLLM-based idea is naturally suited to model the richer visual content from multiple SKU images.
  }
  
  % \vspace{-6mm}
  
  \label{fig:data1}
\end{figure}

To address these limitations, recent years have witnessed the emergence of product-content-based representation learning approaches that leverage the multimodal information of products, including both visual images and textual titles. 
Most of these existing works~\citep{zhan2021product1m,liu2023multimodal,jin2023learning,dong2022m5product,dai2024uniembedding} 
are derived from the dual-flow architecture, employing a visual encoder and a text encoder to separately process multimodal content for each product. 
% While these approaches demonstrate intuitive merit and drive progress in the field, several shortcomings still persist:
% (1) First, the main training strategy of them involves instance-level contrastive learning based on each item's image-text pair or masked modeling, 
% neglecting the positive user feedback (e.g., purchases).
% % which provide richer semantics and enhance the ability of the learned representations to capture correlations between different products. 
% As a result, the learned representations fail to fully capture the inter-product correlations grounded in user interactions.
% (2) More importantly, as shown in Fig.~\ref{fig:data1}, the modeling paradigm of one image corresponding to one title (abbreviated as one-to-one) used in the dual-flow structure 
% fails to directly model the many-to-one relationship between different images of the same product (such as stock keeping unit (SKU) images) and the shared title.
While these approaches demonstrate intuitive merit and drive progress in the field, 
as shown in Fig.~\ref{fig:data1}, they are mainly based on a one-to-one modeling paradigm, which fails to directly model the many-to-one relationship between multiple images of the same product (such as stock-keeping unit (SKU) images) and their shared title.
In contrast, in general multimodal fields, some works have explored multimodal large language models (MLLMs) for multimodal representation learning~\citep{zhang2024gme,lin2024mm}, which offer greater flexibility in modeling richer information from multiple images or texts. 
However, in the e-commerce field, very few have yet used a generative approach for product understanding.

Therefore, inspired by the employment of MLLMs~\citep{liu2024improved,wang2024qwen2,zhang2025sharper} for multimodal representation learning, 
we propose to utilize generative MLLMs for product content understanding in e-commerce scenarios. 
Building upon the powerful vision-language capabilities of MLLMs,
% in general scenarios, 
we aim to enhance their understanding of e-commerce-specific knowledge, to learn task-agnostic multimodal representations of products and thereby support a wide range of downstream tasks in e-commerce, including various cross-modal retrieval, product classification, and attribute prediction. 
% However, learning these domain-adapted multimodal representations with MLLMs presents several key challenges.
However, current research leaves much to be explored
in this direction, primarily due to the following challenges.

First, from the viewpoint of model capability, \textbf{the typical architecture of LLM lacks targeted mechanisms for learning the multimodal and multi-aspect content of product data.}
Conventional LLMs are primarily designed for unimodal textual input and do not incorporate specialized components for multimodal content understanding. 
As a result, the vanilla Transformer~\citep{vaswani2017attention} architecture is insufficient for adaptively modeling multiple modalities~\citep{mustafa2022multimodal}, nor can it specifically capture multi-aspect information  of products such as hierarchical categories and fine-grained attributes. 
Therefore, introducing new modules to better model the multimodal and multi-aspect content poses a challenge for MLLM-based product understanding.

\begin{figure}[t]
  \centering
  \includegraphics[width=\linewidth / 100 * 100]{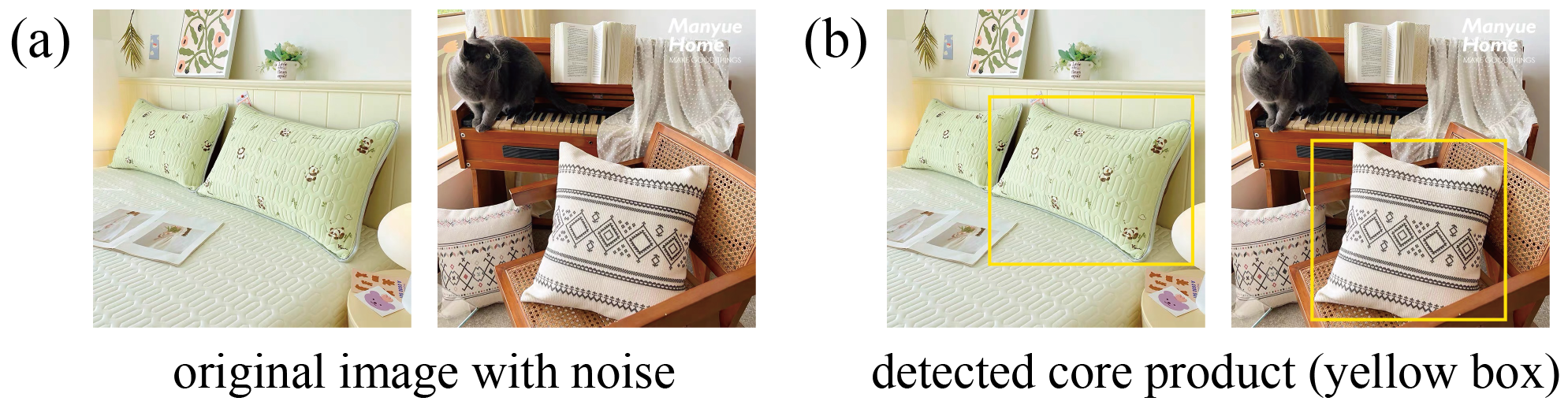}

  % \vspace{-3mm}
  
  \caption{Illustration of the noisy background and the core product. 
  \small 
  (a) Besides the product itself, images often include non-sale objects and background noise. 
  (b) The core semantics of the products for eliminating the distractions of MLLMs.
  }
  
  % \vspace{-6mm}
  
  \label{fig:data2}
\end{figure}

Second, from the perspective of data characteristics, 
\textbf{in addition to the products being sold, product images commonly also contain interference} such as background noise and items not for sale.
For general MLLMs, the ability to exhaustively interpret as many visual details as possible is often seen as a strength~\citep{wang2024qwen2}. 
However, this assumption shows a gap in the e-commerce scenario, where the primary modeling objective is to focus on the product itself—usually indicated by the title—rather than irrelevant noise.
Taking Fig.~\ref{fig:data2}(a) as an example, a product image of a pillow for sale may also include a bed, a chair and other decorations, which are not the item being sold. 
Such extraneous elements can mislead the model and dilute its attention, impairing the understanding of the products.
Therefore, a challenge lies in guiding the model to focus more on the core product within an image while suppressing background noise, in order to mitigate attention distraction and improve the performance of product-centric representation learning.

Finally, from an evaluation perspective, \textbf{existing benchmarks for multimodal general representation in e-commerce are limited in both quantity and quality.} 
% To promote the research in cross-modal product retrieval, 
~\citet{zhan2021product1m} proposes Product1M, an evaluation dataset comprising over 1 million image-caption pairs and fine-grained product categories. 
While Product1M serves as a valuable resource, its product data is restricted to the cosmetics industry, which is far from the data distributions in real-world scenarios,
% This data bias limits its ability to reflect realistic product retrieval challenges.
making it unable to serve the evaluation in actual applications.
To address this, ~\citet{dong2022m5product} proposes M5Product, a multimodal product benchmark supporting multiple downstream tasks. 
However, it only provides product-centric data, excluding any user behaviors, so its retrieval queries can only come from the product images and title, 
lacking the diversity and authenticity of user-generated queries like user-captured photos or colloquial queries. 
Furthermore, it also suffers from the absence of an off-the-shelf evaluation pipeline, lack of hierarchical categories, and the missing modality issues. 
These limitations greatly hinder its effectiveness as a comprehensive benchmark for product understanding.

% To solve these challenges, in this paper, we propose the first MLLM-based \textbf{m}ultimod\textbf{a}l \textbf{r}epresenta\textbf{ti}o\textbf{n} l\textbf{e}arning model for \textbf{pro}duct understanding, named \model, and release a large-scale multimodal benchmark in e-commerce,
% To solve these challenges, in this paper, we propose the first MLLM-based multimodal representation learning model for product understanding, named \model, and release a large-scale multimodal benchmark in e-commerce,
To solve these challenges, 
in this paper, 
we propose the first MLLM-based \textbf{m}ultim\textbf{o}dal 
% representati\textbf{on} learning 
pr\textbf{o}duct u\textbf{n}derstanding model, named \model, and release a large-scale \textbf{m}ultimodal \textbf{b}enchmark for \textbf{e}-commerce, named \bench,
with details as follows:

For the model architecture, we propose a \textit{guided mixture-of-experts (MoE)} technique, which not only enables multiple experts to adaptively model multi-modalities, but also explicitly guides several experts to specialize in learning different aspects of product content,
allowing the model to perform modality-adaptive and aspect-aware representation learning for products. 
Specifically, among the candidate experts, we designate two experts and design a guided routing strategy to direct them to focus on the category and attribute information of products, respectively, enabling the model to capture diverse aspects of product semantics.

For the data augmentation and training strategy, to mitigate the distraction and interference caused by background noise in product images, 
we utilize the visual grounding ability of MLLMs and process the original image into a cropped image of the core product,
% to generate semantically cropped images that focus on the core content of the product. 
so that the model can focus on the item being sold and understand the products more easily (shown in Fig.~\ref{fig:data2}(b)). 
In addition, to better distinguish the fine-grained differences between similar products during contrastive learning, we not only build the hard negative samples for each query, but also expand the pool of negative samples in both spatial and temporal dimensions.
% by employing a historical sample queue and allowing cross-node negative sampling. 
In this way, the MLLM can focus on the core semantics of products and 
% distinguish similar items more effectively.
learn more discriminative representations with more generalizability. 

For evaluation, we publish a large-scale real-world multimodal benchmark named \bench for product understanding, which supports a wide range of downstream tasks, including various cross-modal retrieval, multi-granularity product classification, attribute prediction and so on. 
Our benchmark comprises 2.7M training samples and 410k evaluation samples, all collected from real-world products and user purchases on one of 
% Taobao~\citep{taobao}, 
the largest e-commerce platforms in China. 
The retrieval tasks involved are grounded in actual purchase behaviors rather than trivial category matching, thereby offering a more realistic assessment of the product understanding ability in practical applications. 
Moreover, to establish a unified evaluation standard for product understanding, we provide an off-the-shelf evaluation pipeline to facilitate future research for the community.

To validate the effectiveness of our \model, as shown in Fig.~\ref{fig:radar}, we conduct extensive experiments on both the proposed benchmark and a public dataset, M5Product~\citep{dong2022m5product}. 
The experimental results show that our model achieves state-of-the-art (SOTA) performance in the zero-shot setting, 
demonstrating its capability to generalize across various downstream tasks in e-commerce, including text-based retrieval, image-based retrieval, product-based retrieval, product classification, and attribute prediction. 
Additionally, the visualized attention heatmaps provide compelling evidence that our model captures effective multimodal representations from products, seamlessly combining visual and textual information. 
% This empirical validation underscores the robustness and adaptability of our approach when applied to real-world e-commerce scenarios.
Overall, our key contributions are summarized as follows:

\begin{itemize}[leftmargin=*]
    \item 
    % To our knowledge,
    Beyond the traditional dual-encoder paradigm, 
    we are the first to propose \model, an MLLM-based method for general representation learning of products, 
    which can be applied to multiple downstream tasks, including various cross-modal retrieval, product classification, and attribute prediction.
    \item 
    % Built upon a generative MLLM, we enhance product-level multimodal understanding and modeling by introducing techniques from three key perspectives: **semantic cropping** for data augmentation, **guided expert Mixture-of-Experts (MoE)** for model architecture, and **spatiotemporal negative sampling** for contrastive training.
    From the perspective of module structure, data augmentation, and training strategy, we employ guided MoE, core semantics detection, and spatial and temporal negative sampling techniques, to effectively model the multimodal content of products.
    \item 
    We release a large-scale real-world benchmark, \bench, for multiple downstream tasks of product understanding, which consists of 3.1M data samples and user purchase behaviors,
    % from Taobao, 
    serving as a valuable resource to facilitate future research.
    \item 
    We validate \model through extensive experiments using our \bench and a public dataset on various downstream tasks.
    Moreover, the case study and visualization illustrate the effectiveness of \model in product understanding. 
\end{itemize}

\section{Problem Formulation}\label{sec:prob_form}

In this section, we formalize the definitions of our downstream tasks where the experiments are conducted. 

\vpara{Cross-modal Product Retrieval.}
Given a user query in the form of text, image, or a combination thereof, the goal of the retrieval task is to return a ranked list of candidate products from a large-scale product corpus that is most relevant to the query. 
\begin{definition}
    Let $\mathcal{Q} = \{ q | q\in \{ t_\text{q}, i_\text{q}, (t_\text{q}, i_\text{q}) \} \}$ denote the set of input queries and $\mathcal{P}$ denote the set of products in the gallery, where $t$ and $i$ denote the textual and visual modalities. 
    The aim is to learn an embedding function $f(\cdot)$ that maps the query $q$ and all candidate items $p' \in \mathcal{P}$ to a unified feature space such that the similarity $\text{sim}(f(q), f(p'))$ is maximized when $p'$ corresponds to a product positively associated with query $q$, and minimized otherwise. 
    \begin{equation}
        \hat{p} = \arg\max_{p' \in \mathcal{P}} \text{sim}(f(q), f(p')) ,
    \end{equation}
\end{definition}
The aforementioned query $q$ corresponds to cases of text $t_\text{q}$, image $i_\text{q}$, or image-text $(t_\text{q}, i_\text{q})$, serving scenarios of text-based, image-based, and item-based searching, respectively.

\begin{figure*}[ht]
  \centering
  \includegraphics[width=\linewidth / 10 * 10]{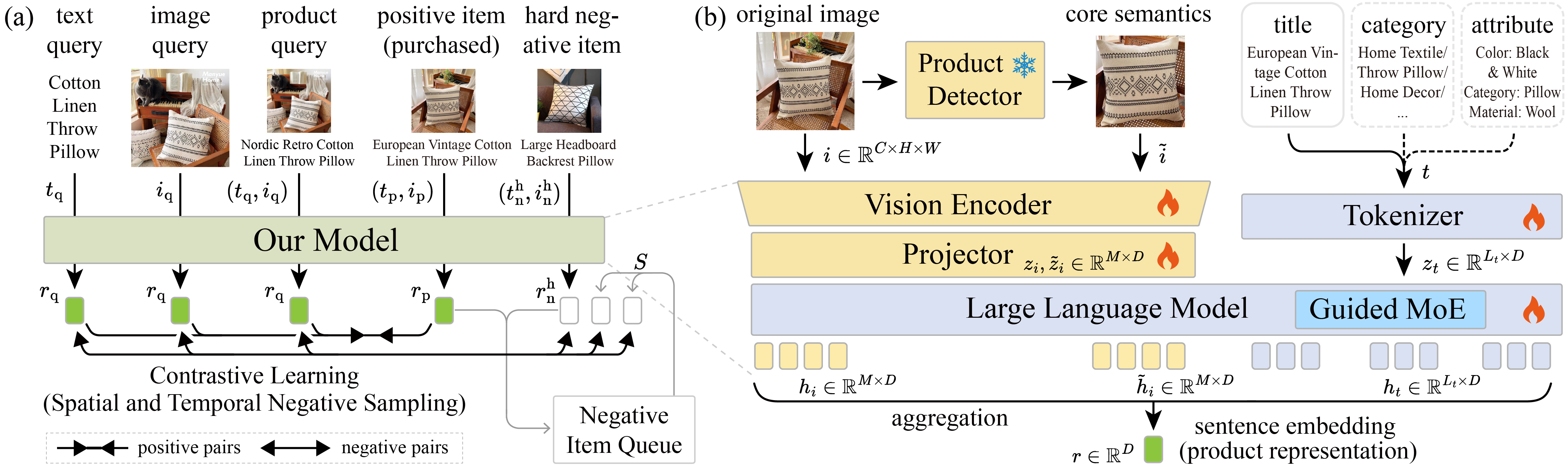}

  % \vspace{-2mm}
  
  \caption{The architecture of our \model.  
  \small 
  % (a) Different from existing contrastive training methods based on the image-text pair of each single product, our supervision signal of contrastive learning are from real-world purchase behaviors, which effectively models the latent correlations between related items. % for various downstream tasks.
  % Moreover, we employ the spatial and temporal negative sampling, fully expanding the negative pool, to learn more generalized and discriminative representations.
  % (b) Moving beyond the traditional dual-encoder paradigm, we propose the first  generative MLLM-based method to learn product representations. First, we employ a product detector to crop the core semantics of the item being sold, which will be fed into the MLLM along with the original image. 
  % To improve the adaptability to multi-modalities and allow aspect-specific modeling of product content, we integrate the guided MoE module into our LLM. 
  % Finally, the mean pooling of the LLM's last hidden states serves as the product representation.
  (a) 
  % Different from existing contrastive learning methods based on image-text pairs of individual products, we 
  We leverage real-world purchase behaviors as supervision to effectively capture latent correlations between related items. 
  Moreover, we employ the spatial and temporal negative sampling, adding hard negative samples and fully expanding the negative pool, to learn more robust and discriminative representations.
  (b) 
  Beyond the dual-encoder paradigm, we propose the first generative MLLM-based method for product understanding. 
  A product detector is used to crop the core semantic region, which is input to the MLLM alongside the original image. 
  % To improve the multimodal adaptability and aspect-aware modeling, we integrate a guided MoE module into the LLM. 
  Finally, the mean pooling of the LLM's last hidden states serves as the product representation.
  }
  
  % \vspace{-1mm}
  
  \label{fig:model}
\end{figure*}

\vpara{Product Classification.}
This task aims to assign each item to one predefined category label.
Following the common practice in CLIP~\citep{radford2021learning}, to unify the task paradigm with that of retrieval, we formulate this task as an embedding-based matching task. 
\begin{definition}
    Given a product item $(t, i)$ and its category $t_\text{cate}$, 
    % where $t$ denotes the text and $i$ is the image, 
    the objective is to learn an embedding function $f(\cdot)$ that maps the product $(t, i)$ and all candidate categories $t_\text{cate}' \in \mathcal{C}$ into a unified feature space, 
    such that the similarity $\text{sim}(f(t, i), f(t_\text{cate}'))$ 
    % $\text{sim}(f((t, i)), f(t_\text{cate}))$ 
    is maximized when $(t, i)$ belongs to $t_\text{cate}'$.
    \begin{equation}
        \hat{t}_\text{cate} = \arg\max_{t_\text{cate}' \in \mathcal{C}} \, \text{sim}(f(t, i), f(t_\text{cate}')) ,
    \end{equation}
\end{definition}
% The candidate textual label of each category is also encoded into the same space. Then, classification is performed by computing the similarity between the product and all candidate category embeddings.

\vpara{Attribute Prediction.}
This task aims to identify the correct attribute values of a given product.
Similarly, we cast attribute prediction as a matching problem between the product and candidate attribute values in a unified representation space.
\begin{definition}
Given a product item $(t, i)$ and its attribute list $\{(t_\text{key},t_\text{val})|t_\text{key} \in \{\text{color,material,}\dots\} \}$, 
the goal is to learn an embedding function $f(\cdot)$ that maps the product $(t, i)$ and all candidate attribute values $t_\text{val}' \in \mathcal{A}$ into a unified space, such that the similarity
is maximized when $(t, i)$ corresponds to the value $t_\text{val}'$.
\begin{equation}
\hat{t}_\text{val} = \arg\max_{t_\text{val}' \in \mathcal{A}} \text{sim}(f((t, i)), f(t_\text{val}') ),
\end{equation}
\end{definition}

% This formulation naturally supports multi-attribute and multi-value scenarios by computing attribute-wise predictions across different candidate sets.

\section{Proposed Method}

In this section, we introduce the technical details of our proposed model, \model.
As shown in Fig.~\ref{fig:model}(a), built upon a pretrained generative vision-language MLLM, our model can be fed with arbitrary input modalities, including text-only, image-only, and image-text queries.
This unified design enables support for  multiple downstream search scenarios. % , including text-based, image-based, and product-based retrieval. 
For the fine-tuning task, we adopt a contrastive learning objective, 
% that pulls together queries and positive samples in the feature space in any modality. 
where the supervision signal is based on the real-world user purchase behavior rather than trivial image-text pairs.
% allowing the learned product representations to better capture user preferences and to naturally accommodate many-to-one relationships such as multiple product images sharing a common title (details in Fig.1).
This not only allows the learned representations to better capture the latent relationship between correlated product items that grounded in user feedback, but also naturally accommodates the information-richer situation where multiple item images correspond to a shared title (details in Fig.~\ref{fig:data1}).
% Moreover, to learn more generalized and discriminative representations in contrastive learning,
Moreover, to learn more robust and discriminative representations 
% to distinguish similar items 
during contrastive training, 
we further employ the spatial and temporal negative sampling mechanism, to introduce hard negative samples and fully expand the pool of the negatives both over time and across distributed nodes (details in Sec.~\ref{sec:neg_samp}).

For the model architecture, as shown in Fig.~\ref{fig:model}(b), moving beyond the traditional dual-encoder paradigm, we propose the generative MLLM-based idea to learn unified product representations. 
Specifically, to ensure that image inputs focus on the product itself rather than irrelevant noise, we adopt the visual grounding capabilities of Qwen2.5-VL~\citep{bai2025qwen25vl} to detect a sub-image of the core semantics. 
Then, both the original image and the cropped core are fed into the vision encoder of MLLM.
For the text, the product title is tokenized and fed into the LLM, which jointly models cross-modal features between the textual and visual embeddings. 
Notably, for various cross-modal retrieval tasks, in addition to the item title, the product category and attributes will also be incorporated to enrich the understanding of more aspects of product contents.
Here we propose the guided mixture-of-experts (MoE) technique within the feed-forward layers, 
to adaptively model multi-modalities and targeted encode different aspects of products.
Finally, as shown in the lower part of Fig.~\ref{fig:model}(b), the token-level embeddings output by LLM are aggregated through mean pooling to yield a sentence-level embedding, which serves as the final representation of the input product for various downstream product understanding tasks.

\vspace{-3mm}

\subsection{Overall Archituecture}

\vpara{Core Product Detection. }\label{sec:core_crop}
As shown in Fig.~\ref{fig:model}(b), given an image of the input item, to mitigate the distraction caused by background noise, we first utilize Qwen2.5-VL~\cite{bai2025qwen25vl} as a detector to detect the bounding box of the core product being sold, as indicated by the title, yielding a cropped image. 
Afterwards, the core image is used alongside the original image for joint modeling within the MLLM.
Formally, the input item is denoted as a tuple $(t, i)$. Here $t$ denotes the textual input, containing the product title, category, and attribute; and $i \in \mathbb{R}^{C \times H \times W}$ denotes the original image, where $C$, $H$, and $W$ denotes the number of channels, height, and width, respectively. 
During training and inference, the input $(t, i)$ may correspond to one of the following: a query $(t_\text{q}, i_\text{q})$, a positive sample $(t_\text{p}, i_\text{p})$, or a negative sample $(t_\text{n}, i_\text{n})$.
Coming from the original image $i$, the detected core image is denoted as $\tilde{i} \in \mathbb{R}^{C \times \tilde{H} \times \tilde{W}}$, where $\tilde{H}$ and $\tilde{W}$ denote its height and width, respectively. 
Both the original image $i$ and the core image $\tilde{i}$ will be fed into the MLLM for task-agnostic multimodal representation learning.

\vpara{Generative-based Embedding. }
% To overcome the limitations of traditional dual-tower architectures in modeling user feedback and many-to-one relationships, we propose a generative MLLM-based method for product representation learning.
Following common practice in the field of MLLM, we feed both the original product image and its core sub-image into the vision encoder and projector of the MLLM to obtain visual embeddings. 
For the text, the product title—may along with category and attributes—is tokenized and jointly processed with the visual embeddings by the LLM for cross-modal joint modeling.
Here our LLM integrates the guided MoE module, to adaptively model different modalities and focus on specific product aspects like category and attribute (details in Sec.~\ref{sec:moe}). 
Finally, the hidden states from the LLM's last layer are aggregated using mean pooling, yielding a sentence-level embedding, which serves as the final representation of the input product.

Formally, the original image $i$ and the cropped image $\tilde{i}$ are passed through the vision encoder and projector to obtain visual embeddings $z_i \in \mathbb{R}^{M \times D}$ and $\tilde{z}_i \in \mathbb{R}^{M \times D}$, where $M$ denotes the number of visual tokens per image, and $D$ is the hidden dimension of the LLM. 
The text input $t$ is embedded as $z_t \in \mathbb{R}^{L_t \times D}$, where $L_t$ is the length of the full text.
Then, the visual embeddings $z_i$, $\tilde{z}_i$, and the textual embeddings $z_t$ are  fed into the LLM. Finally, the LLM's output is the hidden states $h \in \mathbb{R}^{(2M + L_t) \times D}$ from its final layer. We apply mean pooling over $h$ to obtain a sentence embedding $r \in \mathbb{R}^D$, which is the final representation for various downstream tasks.

\begin{figure}[t]
  % \vspace{-3mm}
    
  \centering
  \includegraphics[width=\linewidth / 100 * 100]{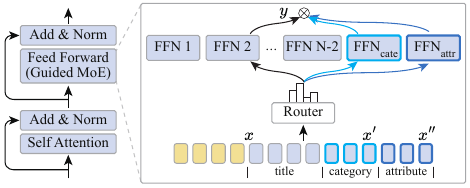}

  % \vspace{-1mm}
  
  \caption{The guided MoE module. 
  \small 
  We replace the FFN layer with a guided MoE module, which explicitly designate two specialized experts to handle the category and attributes within the text input.
  % , improving the adaptability to multi-modalities and allowing aspect-specific modeling of product
  }
  \label{fig:moe}
\end{figure}

\vpara{Guided Mixture-of-Experts.}\label{sec:moe}
Considering that the vanilla Transformer is designed for unimodal text, we propose a guided MoE module to replace the feedforward network (FFN) in the LLM.
% , improving the adaptability to multi-modalities and allowing aspect-specific modeling of product content. 
As shown in Fig.~\ref{fig:moe}, in addition to the basic MoE structure, we explicitly designate two specialized experts to handle category- and attribute-related information within the textual input.
Formally, the input text sequence $t$ is constructed by concatenating the product title, category, and attribute: $
t = \text{Concat}(\text{title}, \text{category}, \text{attribute}).
$
In the Transformer, the intermediate hidden states of the entire sequence are denoted as $x \in \mathbb{R}^{(2M + L_t) \times D}$, where $x = \text{Concat}(x, x', x'')$, and $x'$, $x''$ correspond to the token embeddings associated with the category and attribute substrings, respectively.
We replace each FFN layer with $N$ experts, $\{ E_n \mid n=1, \dots, N \}$, and set a linear layer as a routing function $G(x)$ to compute the weighted combination of expert outputs. 
Specifically, we reserve two dedicated experts, $E' = E_{N-1}$ and $E'' = E_N$, explicitly routing the category tokens $x'$ and attribute tokens $x''$ to these two experts, respectively. 
The final MoE output $y \in \mathbb{R}^{(2M + L_t) \times D}$ is computed as:
\begin{equation}
y = \sum_{n=1}^{N-2} G_n(x) E_n(x) + G_{N-1}(x) E'(x') + G_N(x) E''(x''),
\end{equation}
where the gating weight $G_n(x)$ is defined as 
$
G_n(x) = \frac{\exp(g_n(x))}{\sum_{m=1}^{N} \exp(g_m(x))},
$
and $g(x)$ denotes a linear layer.
Note that if an expert is not activated during inference, its corresponding weight $G$ will be set to 0.
Through this mechanism, our model not only learns to adaptively handle heterogeneous modality inputs, but also explicitly captures multiple aspects of the product content, which significantly improves the ability to learn multimodal product representations for e-commerce scenarios.

\subsection{User Behavior-based Contrastive Learning}
% In the field of product modeling, the supervision signal for training under the traditional dual-tower architectures is typically based on the image-text pair of the same product item (as shown in Fig.~\ref{fig:data1}).
% traditional dual-tower architectures are typically trained using supervision signals derived from paired image-text data of the same product item (as shown in Fig.?).
% That is, the two modality-specific encoders independently encode the visual and textual inputs of an item, and optimizing is performed by pulling together the representations of matching image-text pairs and pushing apart those of different items in the feature space.
% That is,
For the traditional dual-tower methods, two modality-specific encoders independently encode the visual and textual inputs of an item (Fig.~\ref{fig:data1}(a)), which may subsequently be followed by a modality interaction module, and optimization is performed through tasks such as contrastive learning or masked  modeling~\citep{liu2023multimodal,dong2022m5product,zhan2021product1m,jin2023learning}.
However, such an instance-level learning paradigm treats each product in isolation and fails to explicitly model the potential correlations between related items that are implicitly revealed in user interactions,
which are especially important for downstream tasks such as user-oriented product searching. % 这块表述不错 挺好听

To address this limitation, we optimize our contrastive training using supervision signals from positive user feedback, i.e., real-world purchase behaviors. 
Specifically, we mine valuable purchase logs to construct positive training pairs, where an item actually purchased by the user is treated as a positive sample for a given query. 
As shown in Fig.~\ref{fig:model}(a), during training, our model encodes the query $(t_\text{q}, i_\text{q})$ and its corresponding positive item $(t_\text{p}, i_\text{p})$ to yield representations $r_\text{q}$ and $r_\text{p}$, respectively, and they will be pulled closer in the embedding space.
% Formally, given an input query $(t_\text{q}, i_\text{q})$ and the positive item $(t_\text{p}, i_\text{p})$, their representations can be denoted as $r_\text{q}$ and $r_\text{p}$, respectively.
In basic contrastive learning, negative samples are usually drawn from other instances within the same mini-batch. 
However, 
% this strategy does not guarantee high-quality negative samples. 
% In other words, 
considering the vast diversity of product categories, most randomly sampled negatives are trivially dissimilar from the query, which limits the model’s ability to perceive fine-grained distinctions between visually or semantically similar items. 
Therefore, we further introduce a stronger negative sampling strategy to substantially increase the quality of negative candidates, thereby improving the effectiveness of contrastive training, with its details as follows.

\begin{figure}[t]
  \centering
  \includegraphics[width=\linewidth / 100 * 100]{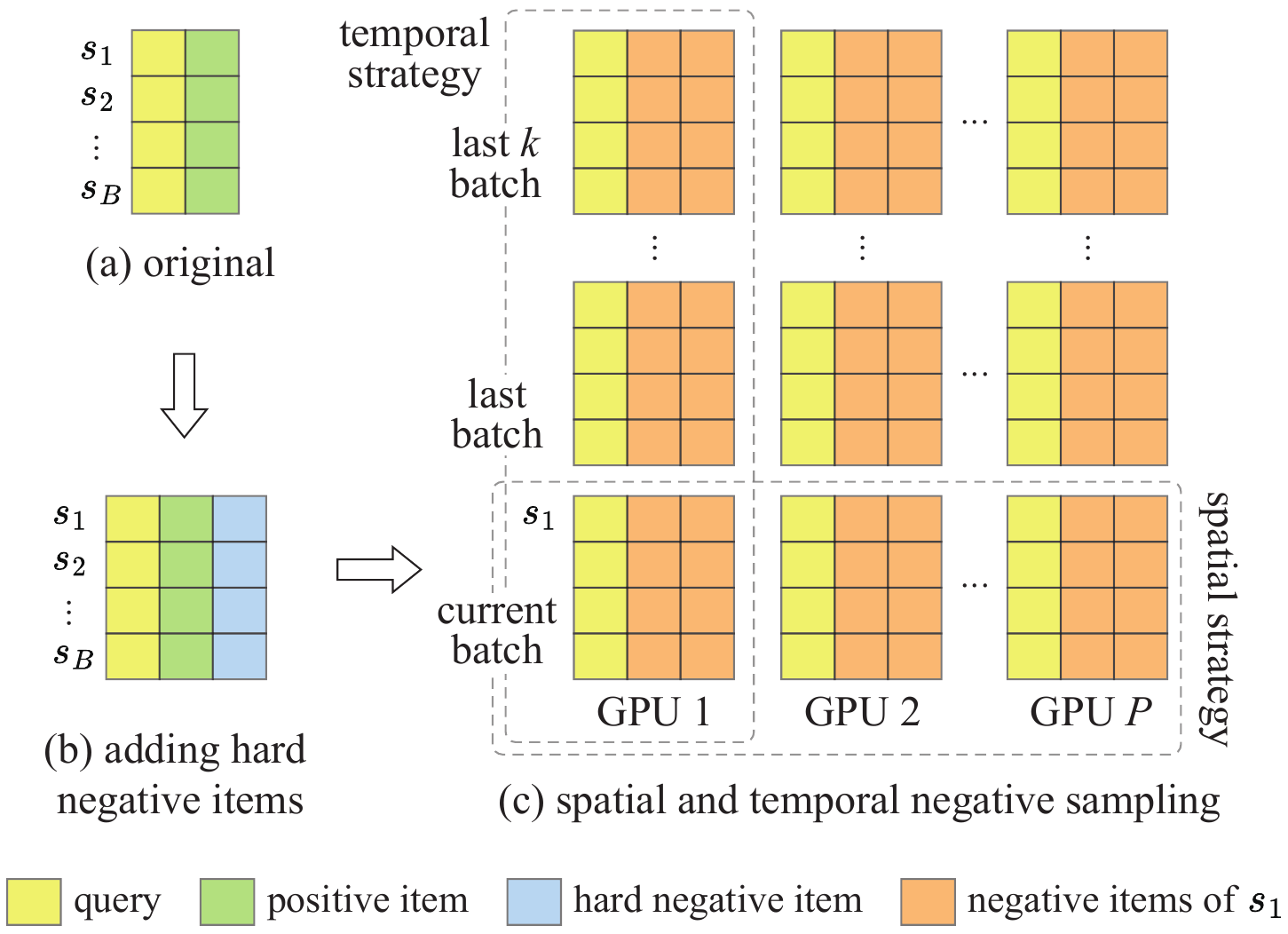}
  
  % \vspace{-2mm}
  
  \caption{Illustration of the spatial and temporal negative sampling. 
  \small 
  (a) For trivial in-batch sampling, negative items are sampled from other items in the same batch.
  (b) We prepare a similar item belonging to the same category as the query as a hard negative item.
  (c) Taking the sample $s_1$ as an example, we greatly expand the negative pool from both spatial and temporal dimensions.
  }

  % \vspace{-1mm}
  
  \label{fig:neg_sampling}
\end{figure}

\vpara{Spatial and Temporal Negative Sampling. }\label{sec:neg_samp}
As discussed above, in order to learn more generalized and discriminative representations, 
during our training, we not only increase the difficulty of distinguishing negative samples, but also expand the sampling range of negative samples beyond the current mini-batch.
As shown in Fig.~\ref{fig:neg_sampling}, we introduce a spatial and temporal negative sampling mechanism~\citep{yan2025mim}. % 自引
Specifically, to introduce harder negatives, in the data preprocessing stage, we prepare a similar product belonging to the same category as the query as a hard negative item $(t^\text{h}_\text{n}, i^\text{h}_\text{n})$ for each query $(t_\text{q}, i_\text{q})$. 
This transforms each training instance from a query-positive pair into a triplet $\{(t_\text{q}, i_\text{q}), (t_\text{p}, i_\text{p}), (t^\text{h}_\text{n}, i^\text{h}_\text{n})\}$, where $(t_\text{p}, i_\text{p})$ denotes the positively purchased item by the user.
Moreover, to further enlarge the pool of negative samples, we adopt the following two strategies:
(1) temporal strategy: negative samples are collected not only from the current batch but also from the past $k$ training batches, increasing the number of negatives to $2B(k + 1) - 1$, where $B$ is the batch size.
(2) spatial strategy: in distributed training, the items from all $P$ GPUs (including all the training nodes) across $k$ recent batches are added as negatives, resulting in $2BP(k + 1) - 1$ total negatives. 
Combined, these strategies provide nearly 200× more negative samples compared to the naive in-batch sampling, significantly improving the model's ability to distinguish subtle differences between semantically similar products in practice.

Finally, we adopt the InfoNCE loss~\citep{oord2019representation} to maximize the mutual information between the query $r_\text{q}$ and its positive item $r_\text{p}$:
\begin{equation}
\mathcal{L} = 
-\log 
\frac{\exp\left(
r_\text{q}^\intercal \cdot r^{}_\text{p} / \tau
\right)}
{
\exp\left(r_\text{q}^\intercal \cdot r^{}_\text{p} / \tau \right) + 
\sum_{r^{'}_\text{n} \in \{r^\text{h}_\text{n}\} \cup S } 
\exp\left(
r_\text{q}^\intercal \cdot r^{'}_\text{n} / \tau
\right)
},
\end{equation}
where $\tau$ denotes the temperature hyperparameter to adjust scale, $r^\text{h}_\text{n}$ denotes the representation of the hard negative $(t^\text{h}_\text{n}, i^\text{h}_\text{n})$, and $S$ denotes the set of additional negatives obtained via spatial and temporal sampling. $\mathcal{L}$ denotes the InfoNCE loss between the query and the user-purchased positive item.

\section{The \bench Benchmark}\label{sec:bench}

\begin{figure*}[t]
  \centering
  \includegraphics[width=\linewidth / 100 * 100]{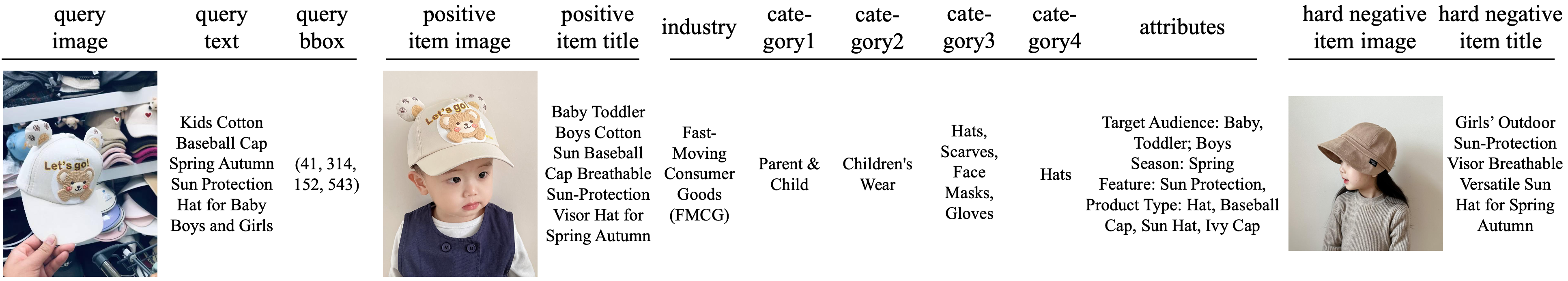}

  % \vspace{-3mm}
  
  \caption{Illustration of one data sample in our benchmark.  
  \small Each data sample includes the images and titles of the query, positive item, and hard negative item, as well as the hierarchical categories and attributes of the positive item, supporting various tasks in e-commerce.
  }

  \vspace{-3mm}
  
  \label{fig:data_sample}
\end{figure*}

\subsection{Collection and Sampling}

% \vpara{Data Collection and Sampling.}
Our data is collected from user logs on one of the largest e-commerce platforms in China. 
Each user purchase interaction is abstracted as a tuple of (user query, purchased item). 
All the data in the training and test set are collected from user behavior data from both text-based and image-based search in real-world scenarios.
The data in our benchmark spans a full year, from January 1, 2024, to December 31, 2024. Given the massive volume of these raw user interactions, we perform sampling to construct a manageable dataset while preserving the original distribution across different industries.
Note that a simple random sampling strategy is insufficient, as certain niche industries, such as digital virtual goods or auctions, constitute less than 0.1\% of the overall traffic, and may be underrepresented or entirely missed. 
To address this, we perform specific sampling based on the original industry-level distribution, ensuring that the sampled data faithfully reflects the real-world distributions of all industries.
The final training and test sets consist of 2,694,397 and 416,926 data samples, respectively.

\subsection{Processing and Format}

% \vpara{Data Processing}
As discussed in Sec.~\ref{sec:neg_samp}, to encourage the model to learn more generalized and discriminative representations during contrastive learning, we construct hard negative samples for each query. 
Specifically, for every query, we sample an item from the same category but of a different product identity as the hard negative. 
This transforms the original (query, purchased positive item) into a triplet (query, purchased positive, hard negative). 
To ensure user privacy, all personally identifiable information is removed; only the visual and textual content of the products is retained.

% \vpara{Data Format.}
An example of a final data sample is illustrated in Fig.~\ref{fig:data_sample}. 
Each sample includes the images and titles of the query, positive item, and hard negative item, as well as the hierarchical category labels and attribute annotations of the positive item. 
This supports a wide range of downstream tasks, including various cross-modal retrieval, multi-granularity product classification, attribute prediction, and so on.
Furthermore, to facilitate further research, we also release the bounding boxes of semantically core regions mentioned in Sec.~\ref{sec:core_crop}, provided under the key ``query bbox'', formatted as $(x_1,y_1,x_2,y_2)$.
Since the original online logs are recorded in Chinese, our data is presented in Chinese. Nonetheless, given that the content primarily consists of discrete phrases and tokens, it is feasible to translate the data into other languages if needed. 
More details about the categories and attributes are given in App.~\ref{app:bench_care_attr}.

\section{Experiment}

\subsection{Experimental Setup}

\vpara{Training. }
Based on our inner-developed generative multimodal large model for the e-commerce field, 
% Based on TBStars~\citep{tbstars}, a generative multimodal large model for the e-commerce domain, 
we perform supervised fine-tuning on our proposed training set (Sec.~\ref{sec:bench}). 
To improve the model capability in handling diverse cross-modal scenarios, we adopt a mixed training strategy involving three types of query modalities: (1) image-only queries, (2) text-only queries, and (3) queries containing both image and text. 
For all cases, the target item always includes both visual and textual modalities. 
These three settings are combined in a ratio of 12:3:2, resulting in an effective training size 1.7× larger than the original dataset, totaling 4,580,475 training samples.
The model is optimized using a learning rate of $1\times 10^{-5}$ with a cosine scheduler and a warmup ratio of 0.05. 
The entire training is conducted on 8 computation nodes and 64 GPUs (NVIDIA H20) with a global batch size of 32, taking about 16 hours.

\vpara{Downstream Tasks. }
As our goal is to learn general representations for e-commerce products, we conduct extensive evaluations across multiple downstream tasks, including text-based retrieval, image-based retrieval, item-based retrieval, fine-grained product classification, and attribute prediction. 
% to assess the model capacity for comprehensive product understanding.
To better demonstrate the generality of the learned representations, all experiments are conducted in a zero-shot setting, without any fine-tuning on the target test distributions. 
In addition to the test set of our \bench benchmark (Sec.~\ref{sec:bench}), we further evaluate the model's generalization ability on the public M5Product~\citep{dong2022m5product} dataset.
More details of these downstream tasks are described in Sec.~\ref{sec:prob_form}. 

For the three cross-modal retrieval tasks, we use $\text{Recall@}k$ as the evaluation metric, which measures the probability that the ground-truth item appears in the $\text{top-}k$ results of the ranked list returned by the model. 
Since all the evaluations on M5Product are conducted in the zero-shot setting, performance at lower $k$ values can be limited for both our model and baselines. Therefore, we report results at $k$ = 10, 100, and 500.
For the product classification task, we adopt standard classification metrics, i.e., accuracy, precision, recall, and F1 score.
% , with further details provided in App.~\ref{}. 
Notably, our classification granularity reaches the fourth level in the category hierarchy (Sec.~\ref{sec:bench}), resulting in a large number of distinct classes. To mitigate sparsity, we relax the criterion: a prediction is considered correct if the true category appears in the model’s top-10 predictions.
For the attribute prediction task, as described in Sec.~\ref{sec:prob_form}, it is also unified under the general representation paradigm and thus evaluated using the same metrics as classification. All evaluation protocols are applied consistently to both our method and all baselines, ensuring a fair comparison across models.

\vpara{Baselines.}
Given our objective of learning general multimodal representations for products, we conduct comprehensive comparisons against advanced multimodal representation learning methods, including classical contrastive-based methods: CLIP~\citep{radford2021learning} and SigLIP2~\citep{tschannen2025siglip}; as well as more recent universal representation models: GME~\citep{zhang2024gme} and MM-Embed~\citep{lin2024mm}.
Also, as our work is the first to employ a generative MLLM for product understanding, we further compare our model with open-source MLLM: InternVL3-2B~\citep{zhu2025internvl3} and Qwen2.5-VL-3B~\citep{bai2025qwen25vl}. 
Since these two models are originally pretrained with next token prediction objectives, 
in addition to the zero-shot setting, 
we further fine-tune them on our training set and report the evaluation results. 
During fine-tuning, we align key experimental settings such as global batch size and learning rate with those used in our model to ensure a fair comparison.
Lastly, to compare our model with domain-specific approaches designed for e-commerce, we include FashionCLIP~\citep{chia2022contrastive} as a baseline. 
% More details of these baselines are provided in App.~\ref{}.

\def\one #1 {\textbf{#1}}
\def\two #1 {\underline{#1}~}
\def\thr #1 {*#1}
\def \s   #1 {\footnotesize $\pm$#1}

\begin{table*}[!t]
    \caption{Zero-shot results of the cross-modal retrieval tasks on our \bench benchmark.}
    % \vspace{-3mm}
    \label{tab:retrieval_bench}
    \centering
    \setlength\tabcolsep{4.2pt}
    \renewcommand{\arraystretch}{0.95}  % n倍行距 
    
\begin{tabular}{lrrrrrrrrr}
\toprule
\multicolumn{1}{c}{\multirow{2}{*}{\diagbox{Methods}{Metrics}}} & \multicolumn{3}{c}{Image-based Retrieval}                & \multicolumn{3}{c}{Text-based Retrieval}                & \multicolumn{3}{c}{Item-based Retrieval}               \\
\cmidrule(lr){2-4} \cmidrule(lr){5-7} \cmidrule(lr){8-10}
                  & \multicolumn{1}{c}{Recall@1} & \multicolumn{1}{c}{Recall@5} & \multicolumn{1}{c}{Recall@10} & \multicolumn{1}{c}{Recall@1} & \multicolumn{1}{c}{Recall@5} & \multicolumn{1}{c}{Recall@10} & \multicolumn{1}{c}{Recall@1} & \multicolumn{1}{c}{Recall@5} & \multicolumn{1}{c}{Recall@10} \\
\midrule
CLIP~\citep{radford2021learning}              & 10.23 & 28.79 & 38.61  & 1.44  & 3.90  & 5.40   & 14.38 & 37.21 & 46.66  \\
SigLIP2~\citep{tschannen2025siglip}           & 15.57 & 46.40 & 59.99  & 3.64  & 10.11 & 14.22  & 13.30 & 36.77 & 48.27  \\
FashionCLIP~\citep{chia2022contrastive}       & 18.96 & 52.35 & 65.32  & 2.78  & 7.53  & 10.32  & 19.81 & 53.21 & 65.42  \\
\midrule
InternVL3-2B~\citep{zhu2025internvl3}          & 2.02  & 5.27  & 7.34   & 0.06  & 0.22  & 0.41   & 3.67  & 8.78  & 11.73  \\
Qwen2.5-VL-3B~\citep{bai2025qwen25vl}          & 13.87 & 36.61 & 46.00  & 0.46  & 1.53  & 2.42   & 15.23 & 40.03 & 50.01  \\
InternVL3-2B~\citep{zhu2025internvl3} (SFT)    & \two26.08 & \two80.14 & \two94.01  & \thr16.04 & \two47.36 & \one60.11  & \two26.02 & \thr79.67 & \thr93.64  \\
Qwen2.5-VL-3B~\citep{bai2025qwen25vl} (SFT)    & \thr25.46 & \thr79.58 & \thr93.70  & 15.07 & 46.07 & \two59.79  & \thr25.64 & \two79.85 & \two93.83  \\
\midrule
GME~\citep{zhang2024gme}                        & 15.76 & 46.22 & 59.60  & \one16.92 & \thr47.22 & 58.84  & 19.79 & 55.57 & 68.61  \\
MM-Embed~\citep{lin2024mm} & 10.01 & 29.17 & 39.34 & 12.83 & 36.25 & 46.42 &  17.89 & 49.37 &  60.77 \\
\midrule
Our \model     & \one26.71 & \one83.66 & \one96.28  & \two16.84 & \one47.46 & \thr59.46  & \one26.76 & \one83.67 & \one96.40  \\
\bottomrule
\end{tabular}

\end{table*}

\def\one #1 {\textbf{#1}}
\def\two #1 {\underline{#1}~}
\def\thr #1 {*#1}
\def \s   #1 {\footnotesize $\pm$#1}

\begin{table*}[h]
    \caption{Zero-shot results of the cross-modal retrieval tasks on the M5Product dataset.}
    % \vspace{-3mm}
    
    \label{tab:retrieval_m5}
    \centering
    \setlength\tabcolsep{1.8pt}
    \renewcommand{\arraystretch}{0.95}  % n倍行距 

\begin{tabular}{lrrrrrrrrr}
\toprule
\multicolumn{1}{c}{\multirow{2}{*}{\diagbox{Methods}{Metrics}}} & \multicolumn{3}{c}{Image-based Retrieval}                    & \multicolumn{3}{c}{Text-based Retrieval}                    & \multicolumn{3}{c}{Item-based Retrieval}                   \\
\cmidrule(lr){2-4} \cmidrule(lr){5-7} \cmidrule(lr){8-10}
                  & \multicolumn{1}{c}{Recall@10} & \multicolumn{1}{c}{Recall@100} & \multicolumn{1}{c}{Recall@500} & \multicolumn{1}{c}{Recall@10} & \multicolumn{1}{c}{Recall@100} & \multicolumn{1}{c}{Recall@500} & \multicolumn{1}{c}{Recall@10} & \multicolumn{1}{c}{Recall@100} & \multicolumn{1}{c}{Recall@500} \\
\midrule
CLIP~\citep{radford2021learning}              & 12.69 & 34.23  & 57.73  & 3.09  & 10.15  & 22.38  & 17.41 & 38.72  & 59.76  \\
SigLIP2~\citep{tschannen2025siglip}           & \thr29.86 & \thr58.37  & \one77.45  & 13.30 & 32.42  & 53.77  & 29.14 & 57.59  & 77.74  \\
FashionCLIP~\citep{chia2022contrastive}       & 23.21 & 49.25  & 69.62  & 6.12  & 18.16  & 33.56  & 25.36 & 49.80  & 68.17  \\
\midrule
InternVL3-2B~\citep{zhu2025internvl3}          & 2.29  & 8.39   & 20.22  & 0.33  & 2.44   & 9.48   & 4.31  & 13.38  & 28.05  \\
Qwen2.5-VL-3B~\citep{bai2025qwen25vl}          & 23.91 & 46.54  & 65.24  & 2.15  & 8.39   & 21.15  & 28.30 & 52.24  & 70.17  \\
InternVL3-2B~\citep{zhu2025internvl3} (SFT)    & \two32.96 & \two58.64  & \thr74.48  & \two44.09 & \two70.44  &  \one84.98 & 41.01  & 68.18  & 82.15  \\
Qwen2.5-VL-3B~\citep{bai2025qwen25vl} (SFT)    & 29.01 & 52.14  & 68.64  & 34.43 & 65.96  & 82.35  & 30.45 & 54.81  & 71.50  \\
\midrule
GME~\citep{zhang2024gme}                & 21.90 & 48.95  & 70.23  & \thr40.96 & \thr68.48  & 84.21  & \two44.05 & \two71.57  & \one86.24  \\
MM-Embed~\citep{lin2024mm}   & 15.40 & 39.74 & 63.38 & 38.78 & 67.44 & \thr84.22 & \thr42.41 & \thr69.48 & \thr84.76 \\
\midrule
Our \model              & \one36.36 & \one59.95  & \two76.28  & \one44.29 & \one70.47  & \two84.60  & \one44.32 & \one71.75 & \two85.32 \\
\bottomrule
\end{tabular}

\end{table*}

\subsection{Experimental Results}

Fig.~\ref{fig:radar} summarizes the overall performance of \model and main baselines across various  tasks. 
% As shown, \model consistently achieves state-of-the-art (SOTA) performance on every task, demonstrating its effectiveness in diverse product understanding scenarios. 
Compared with other baseline methods, \model achieves the best performance on all of the five tasks, illustrating the effectiveness of our method in various e-commerce scenarios. 
Detailed comparisons for each task are discussed in the following paragraphs, where in all the tables we mark values ranking the first (\textbf{v}), second (\underline{v}) and third (*v) in each column. 

The results for cross-modal retrieval tasks are shown in Tab.~\ref{tab:retrieval_bench} and Tab.~\ref{tab:retrieval_m5}. 
\model achieves top rankings in most metrics on both evaluation sets, especially under small $k$ conditions, highlighting its ability to learn highly discriminative and generalizable representations for product understanding. 
The two fine-tuned generative baselines, InternVL3-2B and Qwen2.5-VL-3B, basically rank second and third overall, respectively, indicating the strong potential of generative models in the product domain and supporting our motivation to adopt the generative MLLM. 
However, since these baselines do not incorporate the three key components in \model, their performance lags behind, especially in the condition with a smaller $k$ value, which is more meaningful in application.

As shown in Tab.~\ref{tab:cls_attr}, \model also achieves SOTA performance on product classification and attribute prediction tasks. 
For the classification task, universal representation models like GME and MM-Embed show comparable performance to MLLMs, mainly because they have classification as their primary capability objective.
As for attribute prediction, the fine-tuned InternVL3-2B and Qwen2.5-VL-3B again occupy the second and third positions overall, reinforcing the effectiveness of our generative-based idea. 
While these two may show competitive zero-shot performance on precision, we emphasize that accuracy and F1 are more holistic indicators for these tasks, as they balance both precision and recall.
In summary, our \model delivers consistently strong performance across all five tasks on both our proposed benchmark and the public dataset, validating its effectiveness and generalizability for product understanding.

\subsection{Ablation Study}

\begin{figure}[b]
  \centering
  \includegraphics[width=\linewidth / 100 * 90]{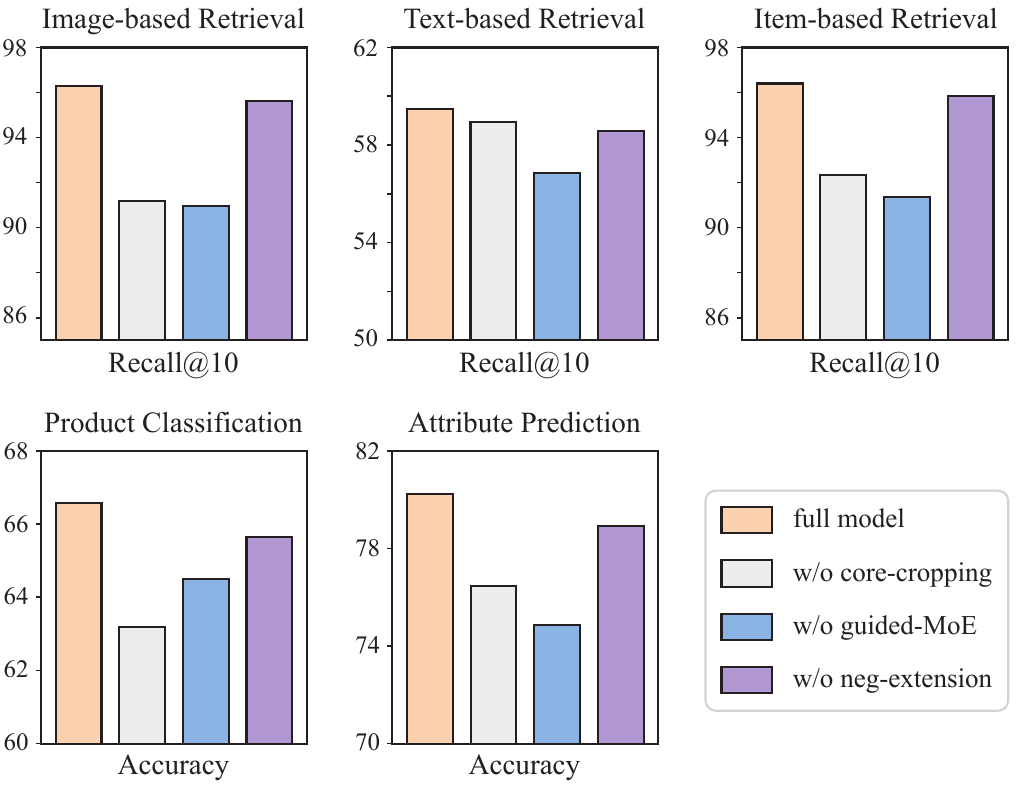}
  % \vspace{-3mm}
  \caption{Results of the ablation study.}
  \label{fig:ablation}
\end{figure}

To assess the contribution of each component in our proposed \model, we conduct ablation experiments across three variants: (1) \textit{\model w/o core-cropping}: the model variant without the detected core region of product images; (2) \textit{\model w/o guided-MoE}: removing the guided MoE module; and (3) \textit{\model w/o neg-extension}: the variant without spatial and temporal negative sampling.
The performance comparison across all five downstream tasks is presented in Fig.~\ref{fig:ablation}. \model consistently outperforms all ablated variants, validating the effectiveness of each component.
Compared with the full model, \textit{\model w/o core-cropping} shows significant performance drops, particularly on tasks that rely heavily on images such as image-based retrieval, highlighting the importance of core product detection in enhancing the model’s understanding of visual semantics. 
Also, \textit{\model w/o guided-MoE} exhibits a notable decline in performance, suggesting that the guided MoE mechanism effectively facilitates targeted modeling of modality-specific and product-specific information. 
Lastly, \textit{\model w/o neg-extension} yields a modest but consistent degradation across tasks, indicating that our spatial and temporal negative sampling strategy helps the model learn more generalized and discriminative representations.

\def\one #1 {\textbf{#1}}
\def\two #1 {\underline{#1}~}
\def\thr #1 {*#1}
\def \s   #1 {\footnotesize $\pm$#1}

\begin{table*}[h]
    \caption{Zero-shot results of the product classification and attribute prediction task.}
    % \vspace{-3mm}
    
    \label{tab:cls_attr}
    \centering
    \renewcommand{\arraystretch}{0.95}  % n倍行距 

\begin{tabular}{
l
*{4}{>{\raggedleft\arraybackslash}p{23pt}}
*{4}{>{\raggedleft\arraybackslash}p{23pt}}
*{4}{>{\raggedleft\arraybackslash}p{23pt}}
}
\toprule
\multicolumn{1}{c}{\multirow{3}{*}{\diagbox{Methods}{Metrics}}}   & \multicolumn{8}{c}{Product Classification}& \multicolumn{4}{c}{\multirow{2}{*}{Attribute Prediction}} \\
\cmidrule(lr){2-9}
   & \multicolumn{4}{c}{Our \bench} & \multicolumn{4}{c}{M5Product} & \multicolumn{4}{c}{}  \\
\cmidrule(lr){2-5} \cmidrule(lr){6-9} \cmidrule(lr){10-13}
   & \multicolumn{1}{c}{Acc.} & \multicolumn{1}{c}{Prec.} & \multicolumn{1}{c}{Rec.} & \multicolumn{1}{c}{F1}& \multicolumn{1}{c}{Acc.} & \multicolumn{1}{c}{Prec.} & \multicolumn{1}{c}{Rec.} & \multicolumn{1}{c}{F1}& \multicolumn{1}{c}{Acc.} & \multicolumn{1}{c}{Prec.} & \multicolumn{1}{c}{Rec.} & \multicolumn{1}{c}{F1} \\
\midrule
CLIP~\citep{radford2021learning}        & 22.76 & 39.61 & 20.30  & 16.63 & 28.89& 44.22 & 27.12  & 24.75 & 49.18 & 55.90  & 49.79   & 42.07  \\
SigLIP2~\citep{tschannen2025siglip}     & 17.85 & 25.83 & 14.58  & 10.91 & 13.22& 19.29 & 12.14  & 8.98  & 43.01 & 44.96  & 43.97   & 36.92  \\
FashionCLIP~\citep{chia2022contrastive} & 44.34 & 53.77 & 48.47  & 41.84 & 41.88& 59.74 & 40.54  & 37.10 & 62.74 & 65.02  & 58.77   & 52.44  \\
\midrule
InternVL3-2B~\citep{zhu2025internvl3}   & 9.16 & 33.31 & 11.52  & 10.60 & 13.36& 44.29 & 13.77  & 12.09 & 38.15 & \two76.27  & 41.89   & 37.28  \\
Qwen2.5-VL-3B~\citep{bai2025qwen25vl}   & 35.64 & \thr62.44 & 35.26  & 36.29 & 51.62& 78.06 & 48.94  & 45.74 & 58.19 & \one80.20  & 54.98   & 51.79  \\
InternVL3-2B~\citep{zhu2025internvl3} (SFT) & 53.12 & 58.09 & \thr63.96  & 51.21 & 81.13 & 87.02 & 83.74  & 81.83 &  \two74.07 & 68.13  &  \two77.03 & \two69.58 \\
Qwen2.5-VL-3B~\citep{bai2025qwen25vl} (SFT) & 55.50 & \two63.47 & \two64.75  & 53.56 & \two88.82 & \thr89.45 & \two88.23  & \two87.43 & \thr72.20 & 68.60  & \thr74.89   & \thr67.56  \\
\midrule
GME~\citep{zhang2024gme}    & \thr57.06 & 59.14 & 62.31  & \thr54.16 & 80.76 & 83.96 & 81.09  & 79.98 & 70.17 & 68.46  & 72.44   & 65.80  \\
MM-Embed~\citep{lin2024mm}  & \two57.56 & \one71.55 & 63.91 & \two57.89 & \thr88.40 & \two91.00 & \thr87.65 & \thr86.79 & 61.04 & 68.04 & 62.71 & 56.18 \\
\midrule
Our \model   & \one66.57 & 62.24 & \one71.26  & \one63.19 & \one92.92 & \one93.09 & \one93.10  & \one92.48 & \one80.22 & \thr74.67  & \one81.90   & \one75.49 \\
\bottomrule
\end{tabular}
\end{table*}

\subsection{Case study}

\begin{figure}[b]
  % \vspace{-3mm}
  \centering
  \includegraphics[width=\linewidth / 100 * 100]{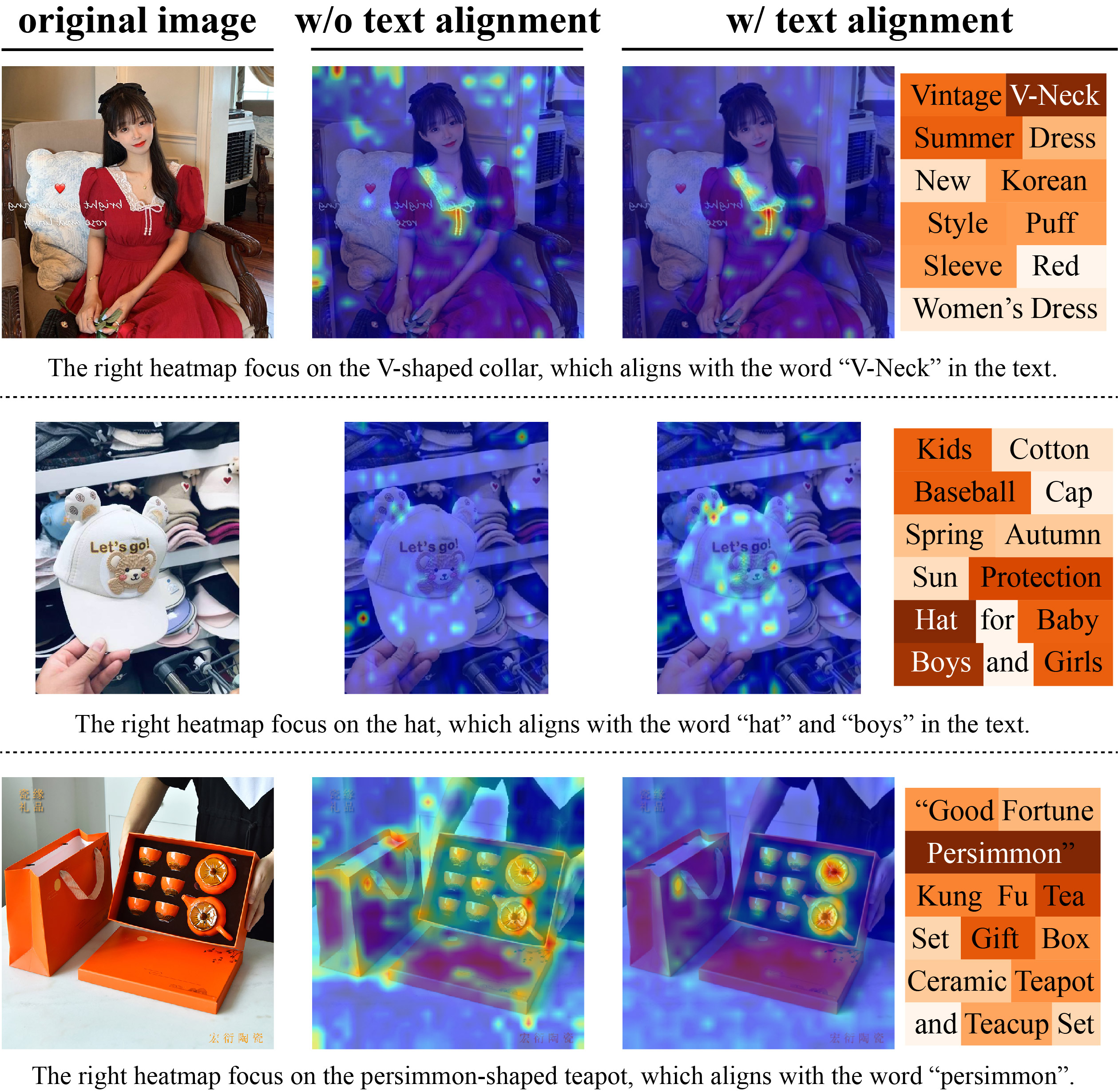}
  % \vspace{-6mm}
  \caption{Visualization of the cases.
  \small For images and texts, the closer the region's color is to red, the higher the model's attention.
  }
  \label{fig:case}
\end{figure}

To more intuitively demonstrate the cross-modal alignment capabilities of our model, we visualize attention heatmaps on the input image (and text) under two conditions: with and without text input. 
As shown in several examples in Fig.~\ref{fig:case}, the heatmaps on the left tend to exhibit dispersed attention, often focusing insufficiently on the core product regions or being distracted by background elements.
In contrast, when textual input is provided, the model is able to attend simultaneously to key visual regions and relevant textual tokens. 
The attended areas in both modalities are semantically aligned, indicating that the model successfully maps image and text inputs into a shared feature space, which highlights not only the alignment ability of our model but also its interpretability in cross-modal product understanding.

\section{Related Work}

\vpara{Multimodal Product Understanding. }
The domain of product understanding encompasses a wide range of product-centric tasks, such as cross-modal product retrieval~\citep{jin2023learning}, product classification~\citep{zhao2018categorical}, attribute prediction~\citep{bai2022commerce,yang2022mave}, and clustering~\citep{dong2022m5product}. 
% These tasks play a crucial role in real-world e-commerce systems, greatly improving user experience for both buyers and sellers, while also contributing to the broader socio-economic development.
FashionBERT~\citep{gao2020fashionbert} was the first work for multimodal understanding in the e-commerce fashion field, learning high-level representations of product images and textual descriptions using pretrained ResNet-50~\citep{he2016deep} and BERT~\citep{devlin2019bert} encoders.
Building upon CLIP, \citet{chia2022contrastive} propose FashionCLIP, a contrastive learning model for fashion industry, demonstrating strong generalization across diverse tasks  and datasets.
\citet{yu2022commercemm} propose CommerceMM, a multimodal framework for commerce understanding, incorporating five image-text pair tasks and an additional nine cross-modal and cross-pair retrieval tasks to facilitate effective pretraining.
\citet{jin2023learning} introduce an instance-centric multimodal pretraining paradigm aimed at constructing a general-purpose foundation model.
\citet{liu2023multimodal} pretrained MBSD, a vision-language model designed to integrate heterogeneous item information using a lightweight convolutional backbone for image encoding, thereby reducing computational overhead.
% For recommendation systems, **XXX et al.** developed *UniEmbedding*, a pretraining framework that integrates a domain-aware adapter, user-view projection module, and cross-domain contrastive objectives.
\citet{dai2024uniembedding} propose the UniEmbedding pretraining framework for recommendation, which includes a domain-aware adapter, a user-view projection, and contrastive learning objectives across domains.
Despite these advances, existing methods are mainly built on a dual-encoder architecture and trained with item-centric contrastive learning or masked language modeling objectives. 
As a result, they inherently struggle to handle the natural many-to-one relationship between a product's multiple images (e.g., SKUs) and the title, failing to leverage real-world user feedback signals during representation learning (details in Sec.~\ref{sec:intro}).
To address these limitations, inspired by the advancement of MLLMs~\citep{liu2024improved,wang2024qwen2,bai2025qwen25vl,zhang2025sharper,zhao2025swift}, we propose the first generative MLLM-based pretraining method for product representation learning, 
which not only naturally allows joint learning of richer information from multi-SKU images, but also 
% leverages users’ real-world purchasing behavior as a supervisory signal for contrastive learning.
% Our approach inherently models the richer visual diversity of multi-SKU product imagery and 
directly incorporates user purchase behavior as supervision, bridging the gap between real-world user intent and product understanding.

\vpara{Datasets for Product Understanding.}
To support the rapid iteration of approaches in product understanding, a growing number of e-commerce datasets have been proposed, serving as valuable resources for the research community. 
Early efforts primarily focused on designing task-specific models tailored to distinct e-commerce scenarios, such as product classification~\citep{rostamzadeh2018fashion,liu2023mep}, product captioning~\citep{yang2020fashion}, product retrieval~\cite{zhan2021product1m,dodds2022training,chen2023real20m}, product matching~\citep{bonab2021cross}, machine translation~\citep{song2021product} and so on. 
% However, with the development of e-commerce research, such narrowly scoped methods have revealed critical limitations, including poor generalizability and high overall training costs.
% As a result, there has been a paradigm shift toward universal product representation learning, which necessitates the emergence of datasets and benchmarks for multi-task evaluation.
As e-commerce research advances, task-specific methods show limited generalizability and high training costs, prompting a shift toward universal product representation and multi-task benchmarks.
A representative example in this direction is M5Product~\citep{dong2022m5product}, a multimodal product benchmark comprising five modalities and supporting a range of downstream tasks, including cross-modal retrieval and product classification. 
% Despite its contributions, M5Product suffers from several key limitations. 
% Most notably, 
However, M5Product only contains product information while entirely neglecting user interactions, so retrieval queries are constrained to image-text pairs derived from the same product category, which fail to capture real-world user queries
% —such as photos taken by users or free-form search text—
and omit valuable positive feedback that contains latent product correlations. 
Also, it lacks an off-the-shelf evaluation pipeline, suffers from the absence of hierarchical categories, and exhibits missing modality issues, all of which hinder its effectiveness as a universal multimodal benchmark for e-commerce.
To address this issue, we release a large-scale multimodal benchmark \bench to support a wide range of downstream tasks across various e-commerce scenarios. 
All retrieval tasks in our benchmark are grounded in real-world purchase behavior rather than weakly correlated category-level supervision, offering a more application-driven assessment for product understanding. 
% Furthermore, we provide an off-the-shelf evaluation pipeline to facilitate reproducibility and standardized evaluation.

\section{Conclusion}

Breaking from the conventional dual-encoder paradigm, we are the first to propose a generative MLLM-based method named \model for product content understanding, which can be applied to a wide range of e-commerce tasks, including cross-modal retrieval, product classification, and attribute prediction. 
From architectural design to data augmentation and training strategy, our approach incorporates guided MoE, core semantic detection, and spatial-temporal negative sampling to effectively model the rich multimodal content of products.
Moreover, we release a large-scale real-world benchmark, \bench, for multiple downstream tasks of product understanding, which consists of 3.1M high-quality data samples and user purchase behaviors,  
serving as a valuable resource to facilitate the development of e-commerce applications by the broader research community.
% not only for the evaluation of \model but also for advancing the development of e-commerce applications by the broader research community.
Overall, our work paves a new avenue for generative MLLM-based approaches in product understanding. 
More explorations in this direction are given in~\citet{fu2025moonembedding}~\citep{nie2025moon2}.
% Overall, our work establishes a significant milestone in the field of generative MLLM-based methods for product content understanding. 
% By demonstrating the effectiveness of \model, we pave a new avenue for MLLM-based task-agnostic product representations  for e-commerce scenarios. 
Encouraged by the positive results of our \model, future research can build on our foundation to further explore unified product representation for e-commerce scenarios.

% \clearpage
\appendix

\section{Hierarchical Categories and Attributes in \bench}\label{app:bench_care_attr} 

To our knowledge, our \bench benchmark is the first multi-task product understanding benchmark that provides hierarchical category annotations. 
As shown in Fig.~\ref{fig:data_sample}, each positive item is labeled with a five-level key-value structure, representing the industry name and category levels 1 through 4. 
This supports flexible evaluation across multiple levels of granularity in product classification, allowing researchers to tailor the task difficulty according to specific research objectives.
The number of categories in these five levels are 24, 128, 1163, 2327, and 87518, respectively.
For the attribute information, due to the lack of a standardized schema for attribute keys across all products, we conduct a frequency analysis over the full dataset and 
summarize 10 attributes with high frequency and more research meaning:
% identify ten high-frequency and semantically meaningful attributes: 
color, category, category modifier, brand, size/specification, material, style elements, fashion style, target demographic, and season/occasion. 
% Note that not all products are annotated with all ten attributes. Therefore, the attribute dictionary may contain missing values, and attribute prediction should only be performed over the attributes that are actually present for a given product.
Note that there are empty values in this key-value dictionary because there are almost no attribute keys that are involved in all kinds of products, 
and thus the attribute prediction task is supposed to only be performed over the attributes that are actually present for a given product item.

\section{Evaluation Metrics} \label{app:metrics} 

For the product classification task, we use accuracy, precision, recall, and F1 score as evaluation metrics. 
Detailed information of these metrics is given as follows: 
\begin{itemize}[leftmargin=*]
    \item 
    Accuracy:
    Accuracy is the overall correctness of the model in predicting both positive and negative outcomes. It is the ratio of correctly predicted observations to the total number of observations. While useful in balanced datasets, it can be misleading in imbalanced datasets where one class dominates.
    \begin{equation}
    \text{Accuracy} = \frac{TP + TN}{TP + TN + FP + FN} ,
    \end{equation}
    where TP is true positives, TN is true negatives, FP is false positives, and FN is false negatives.

    \item 
    Precision: 
    Precision is the proportional accuracy of correctly identified positive outcomes out of all predicted positive outcomes. It's a crucial metric when the cost of a false positive is high. The higher the value, the more relevant the results returned by the model. A lower value would mean that the model returns more false positives. 
    \begin{equation}
        \text{Precision} = \frac{TP}{TP+FP} ,
    \end{equation}
    where TP is the number of true positives and FP is the number of false positives.

    \item Recall:
    Recall, also known as sensitivity or true positive rate, measures the proportion of actual positive cases that were correctly identified by the model. It is particularly important when missing positive cases (false negatives) is costly, such as in medical diagnosis. A high recall means the model captures most of the true positives.
    \begin{equation}
    \text{Recall} = \frac{TP}{TP + FN} ,
    \end{equation}
    where TP is the number of true positives and FN is the number of false negatives.

    \item
    $\text{F}_\beta$-score:
    $\text{F}_\beta$ is the weighted harmonic mean of precision and recall, allowing the user to adjust the relative importance of recall versus precision through the $\beta$ parameter. A higher $\beta$ gives more weight to recall, while a lower $\beta$ favors precision. This metric is useful when there's a need to balance or prioritize between false positives and false negatives.
    \begin{equation}
    F_\beta = \frac{ (1 + \beta^2) \cdot \text{Precision} \cdot \text{Recall}}{\beta^2 \cdot \text{Precision} + \text{Recall}} ,
    \end{equation}
    where $\beta$ determines the weight of recall in the combined score. Common choices include $\text{F}_1$ (equal weight) or $\text{F}_2$ (recall is more important).

\end{itemize}

\clearpage
\bibliographystyle{ACM-Reference-Format}
\bibliography{reference}

\end{document}